\documentclass{article} 
\usepackage{iclr2019_conference,times}

\usepackage{amsmath,amsfonts,bm}

\def\eqref#1{equation~\ref{#1}}

\def\1{\bm{1}}

\DeclareMathAlphabet{\mathsfit}{\encodingdefault}{\sfdefault}{m}{sl}
\SetMathAlphabet{\mathsfit}{bold}{\encodingdefault}{\sfdefault}{bx}{n}

\usepackage{hyperref}
\usepackage{url}
\newcommand{\ignore}[1]{}

\newcommand{\prop}{\textsc{ProPara}\xspace}
\newcommand{\prostruct}{\textsc{ProStruct}\xspace}
\newcommand{\proglobal}{\textsc{ProGlobal}\xspace}
\newcommand{\prolocal}{\textsc{ProLocal}\xspace}
\newcommand{\drqa}{\textsc{DrQA}\xspace}
\newcommand{\recipes}{\textsc{Recipes}\xspace}
\newcommand{\entnet}{\textsc{EntNet}\xspace}
\newcommand{\alg}{\textsc{Kg-Mrc}\xspace}

\usepackage{wrapfig}

\usepackage{pslatex}
\usepackage{booktabs}
\usepackage{latexsym}
\usepackage{color}
\usepackage{mathtools}
\usepackage{graphicx}
\usepackage{amsmath}
\usepackage{xspace}
\usepackage{tikz-dependency}
\usepackage{balance}
\usepackage{enumitem}
\usepackage{mathtools}
\usepackage{varwidth}
\usepackage{pifont}
\usepackage{float}
\usepackage{framed} 
\usepackage{verbatim}
\usepackage[rightcaption]{sidecap}
\usepackage{subfig}
\usepackage{wrapfig}
\usepackage{multirow}
\usepackage{graphicx}
\usepackage[font={small}]{caption}

\title{Building Dynamic Knowledge Graphs from Text using Machine Reading Comprehension}

\author{Rajarshi Das\thanks{Work performed when author was an intern at MSR Montr\'{e}al.} $^{1}$, Tsendsuren Munkhdalai$^{2}$, Xingdi Yuan$^{2}$, Adam Trischler$^{2}$, Andrew McCallum$^{1}$ \\
$^{1}$College of Information and Computer Sciences\\
University of Massachusetts, Amherst\\
\texttt{\{rajarshi, mccallum\}@cs.umass.edu} \\
$^{2}$Microsoft Research Montr\'{e}al\\
Montr\'{e}al, Qu\'{e}bec, Canada\\
\texttt{\{tsendsuren.munkhdalai,eric.yuan,}
\texttt{adam.trischler\}@microsoft.com}
}

\iclrfinalcopy 
\begin{document}

\maketitle

\begin{abstract}
We propose a neural machine-reading model that constructs dynamic knowledge graphs from procedural text.
It builds these graphs recurrently for each step of the described procedure, and uses them to track the evolving states of participant entities.
We harness and extend a recently proposed machine reading comprehension (MRC) model to query for entity states, since these states are generally communicated in spans of text and MRC models perform well in extracting entity-centric spans.
The explicit, structured, and evolving knowledge graph representations that our model constructs can be used in downstream question answering tasks to improve machine comprehension of text, as we demonstrate empirically.
On two comprehension tasks from the recently proposed \prop dataset \citep{dalvi2018tracking}, our model achieves state-of-the-art results.
We further show that our model is competitive on the \recipes dataset \citep{kiddon2015mise}, suggesting it may be generally applicable.
We present some evidence that the model's knowledge graphs help it to impose commonsense constraints on its predictions.
\end{abstract}

\section{Introduction}
\label{sec:intro}

Automatically building knowledge graphs (KGs) from text is a long-standing goal in artificial intelligence research. KGs organize raw information in a structured form, capturing relationships (labeled edges) between entities (nodes). They enable automated reasoning, e.g., the ability to infer unobserved facts from observed evidence and to make logical ``hops,'' and render data amenable to decades of work in graph analysis.

\begin{SCfigure*}
\small
\centering
	\includegraphics[width=0.5\columnwidth]{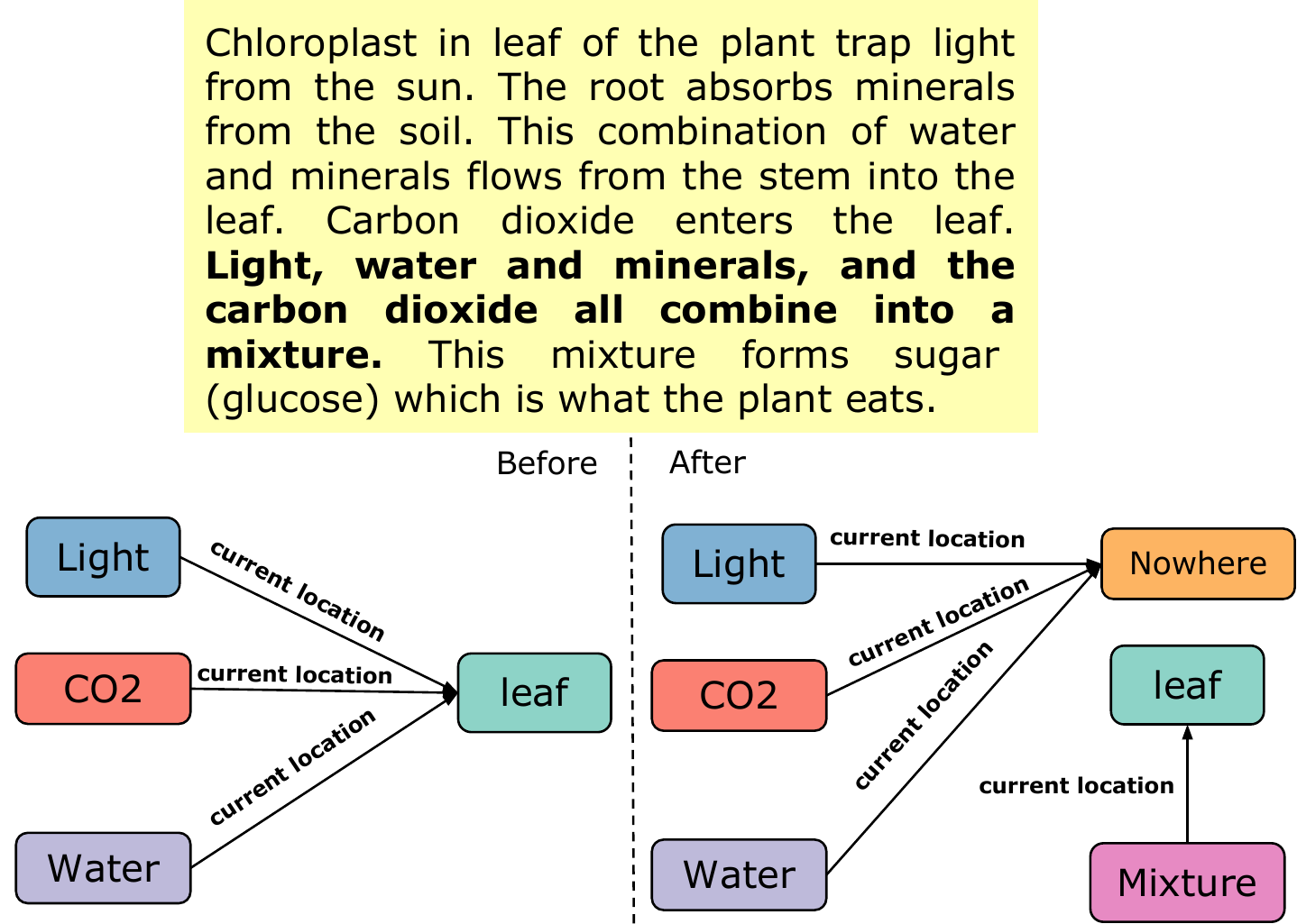}
\caption{Snapshot of the knowledge graphs created by our model before and after reading the sentence in boldface. Since the KG explicitly stores the current location of \textit{light}, \textit{CO$_2$}, and \textit{water} as \textit{leaf}, the model can infer that \textit{mixture} is formed in the \textit{leaf} even though this is not explicitly stated. The three participant entities also get destroyed in the process, which is captured in the graph by pointing to a special \textit{Nowhere} node.}
\label{fig:dynamic_kbs_1}
\end{SCfigure*}

There exists a profusion of text that describes complex, dynamic worlds in which entities' relationships evolve through time. This includes news articles, scientific manuals, and procedural text (e.g., recipes, how-to guides, and so on). Building KGs from this data would not only help us to study the changing relations among participant entities, but also to make implicit information more explicit. For example, the graphs at each step in Figure~\ref{fig:dynamic_kbs_1} help us to infer that the new entity \textit{mixture} is created in the \textit{leaf}, since the previous location of its participant entities (\textit{light, CO$_2$, water}) was \textit{leaf} -- even though this is never stated in the text.

This paper introduces a neural machine-reading model, \alg, that (i) explicitly constructs dynamic knowledge graphs to track state changes in procedural text and (ii) conditions on its own constructed knowledge graphs to improve downstream question answering on the text.
Our dynamic graph model is recurrent, that is, the graph at each time step depends on the state of the graph at the previous time step.
The constructed graphs are parameterized by real-valued embeddings for each node that change through time.

In text, entities and their states (e.g., their locations) are given by spans of words.
Because of the variety of natural language, the same entity/state may be described with several surface forms.
To address the challenge of entity/state recognition, our model uses a machine reading comprehension (MRC) mechanism \cite[inter alia]{seo2016bidirectional,xiong2016dynamic,chen2017reading,yu2018qanet}, which queries for entities and their states at each time step.
We leverage MRC mechanisms because they have proven adept at extracting text spans that answer entity-centric questions \citep{Levy2017ZeroShotRE}.
However, such models are static by design, returning the same answer for the same query and context. Since we expect answers about entity states to change over the course of the text, our model's MRC component conditions on the evolving graph at the current time step (this graph captures the instantaneous states of entities).

To address the challenge of aliased text mentions, our model performs \emph{soft co-reference} as it updates the graph.
Instead of adding an alias node, like \textit{the leaf} or \textit{leaves} as aliases for \textit{leaf}, the graph update procedure soft-attends~\citep{bahdanau2014neural} over all nodes at the previous time step and performs a gated update \citep{cho2014learning,chung2014empirical} of the current embeddings with the previous ones. This ensures that state information is preserved and propagated across time steps.
Soft co-reference can also handle the case that entity states \emph{do not} change across time steps, by applying a near-null update to the existing state node rather than duplicating it.

\ignore{It is also possible that the state of few entities \emph{do not} change at a time step. For example, consider the sentence (in bold) in figure~\ref{fig:soft_coref}.
At this time-step (sentence), only the location of \textit{water} and \textit{minerals} change (to \textit{leaf}) and the location of \textit{light} remains unchanged (as \textit{leaf}). Since there already exists a node corresponding to \textit{leaf} in the graph, instead of replacing the current \textit{leaf} node, our model performs `\textit{soft co-reference}' (\S\ref{sub:coref}) by attending to the locations of entities at the previous time step. Co-reference with soft attention \citep{bahdanau2014neural,luong2015effective} also provides other advantages. For example, the reading comprehension model can return different span for the same surface form (e.g. different mentions of \textit{leaf} in the first and third sentence). Soft attention could still identify that they are the same spans by learning high attention weights between them. It is also robust when the MRC model returns similar spans (e.g. \textit{the leaf} instead of \textit{leaf}).}

At each time step, after the graph has been updated with the (possibly) new states of all entities, our model updates each entity representation with information about its state. The updated information about each individual entity is further propagated to all other entities (\S~\ref{sub:graph_update}).
This enables the model to recognize, for example, that entities are present in the same location (e.g., \textit{light, CO$_2$} and \textit{water} in Figure~\ref{fig:dynamic_kbs_1}).
Thus, our model can use the information encoded in its internal knowledge graphs for a more comprehensive understanding of the text.
We will demonstrate this experimentally by tackling comprehension tasks from the the recently released \prop and \recipes datasets.

Our complete machine reading model, which both builds and leverages dynamic knowledge graphs, can be trained end-to-end using only the loss from its MRC component; i.e., the negative log-likelihood that the MRC component assigns to the span that correctly describes each entity's queried state.
We evaluate our model (\alg) on the above two \prop tasks and find that the same model significantly outperforms the previous state of the art. For example, \alg obtains a \emph{9.92}\% relative improvement on the hard task of predicting at which time-step an entity moves. Similarly on the latter task, \alg obtains a \emph{5.7}\% relative improvement over \prostruct and \emph{41}\% relative improvement over other entity-centric models such as \entnet \citep{henaff2016tracking}.
On the \ignore{related but much harder }\recipes dataset, the same model obtains competitive performance.


\section{Related Work}
\label{sec:related_work}
There are few datasets that address the challenging problem of tracking entity state changes. The bAbI dataset \citep{weston2015towards} includes questions about movement of entities; however, its language is generated synthetically over a small lexicon, and hence models trained on bAbI often do not generalize well when tested on real-world data.  For example, state-of-the-art models like \entnet \citep{henaff2016tracking} and Query Reduction Networks \citep{seo2016query} fail to perform well on \prop.

\textsc{ProRead} \citep{berant2014modeling} introduced the \textsc{ProcessBank} dataset, which contains paragraphs of procedural text as in \prop. However, this earlier task involves mining arguments and relations from events, not tracking the dynamic state changes of entities.
The model that \citet{berant2014modeling} propose builds small knowledge graphs from the text, but they are not dynamic in nature. The model also relies on densely annotated process structure for training, demanding curation by domain experts. On the other hand, our model, \alg, learns to build dynamic KGs just from annotations of text spans, which are much easier to collect.

For the sentence-level \prop task they propose, \citet{dalvi2018tracking} introduce two models: \prolocal and \proglobal. \prolocal makes local predictions about entities by considering just the current sentence. This is followed by some heuristic/rule-based answer propagation.
\proglobal considers a broader context (previous sentences) and also includes the previous state of entities by considering the probability distribution over paragraph tokens in the previous step.
\citet{tandon2018reasoning} recently proposed a neural structured-prediction model, (\prostruct), where hard and soft common-sense constraints are injected to steer their model away from globally incoherent predictions.
We evaluate \alg on the two \prop tasks proposed by \citet{dalvi2018tracking} and \citet{tandon2018reasoning}, respectively,
and find that our single model outperforms each of the above models on their respective tasks of focus.

\entnet \citep{henaff2016tracking} and query reduction networks (\textsc{QRN}) \citep{seo2016query} are two state-of-the-art entity-centric models for the bAbI dataset. \entnet maintains a dynamic memory of hidden states with a gated update to the memory slots at each step. Memory slots can be tied to specific entities, but unlike our model, \entnet does not maintain separate embeddings of individual states (e.g., current locations); it also does not perform explicit co-reference updates. \textsc{QRN} refines the query vector as it processes each subsequent sentence until the query points to the answer, but does not maintain explicit representations of entity states. Neural Process Networks (NPN) \citep{bosselut2017recipe} learn to understand procedural text by explicitly parameterizing actions and composing them with entities.
These three models return an answer by predicting a vocabulary item in a multi-class classification setup, while in our work we predict spans of text directly from the paragraph.

MRC models have been used previously for extracting the argument of knowledge base (KB) relations, by associating one or more natural language questions with each relation (\emph{querification}). These models have been shown to perform well in a zero-shot setting, i.e., for a previously unseen relation type \citep{Levy2017ZeroShotRE}, and for extracting entities that belong to non-standard types \citep{roth2018neural}.
These recent positive results motivate our use of an MRC component in \alg.


\section{Data \& Tasks}
\label{sec:data}
We evaluate \alg on the recently released \prop dataset \citep{dalvi2018tracking}, which comprises procedural text about scientific processes. 
The location states of participant entities at each time step (sentence) in these processes are labeled by human annotators, and the names of participant entities are given.
As an example, for a process describing photosynthesis, the participant entities provided are: \textit{light, CO$_2$, water, mixture} and \textit{glucose}. Although participant entities are thus known \emph{a priori}, the location of an entity could be \emph{any} arbitrary span in the process text.
This makes the task of determining and tracking an entity's changing location quite challenging.
\begin{wraptable}{r}{0.35\textwidth}
\vspace{-8pt}
\centering
\footnotesize
\begin{tabular}{l  c}\hline
    \# para &  488\\
    \# train/\#dev/\#test & 391/43/54\\ 
    avg. \# entities & 4.17 \\
    avg. \# sentences & 6.7 \\
    \# sentences & 3.3K\\\hline
\end{tabular}
\caption{Statistics of \prop.}
\vspace{-12pt}
\label{tab:stats}
\end{wraptable}
It should also be noted that the dataset does not provide information on whether a particular entity is an input to or output of a process. Not all entities exist from the beginning of the process (e.g. \textit{glucose}) and not all exist at the end (e.g. \textit{water}). Table~\ref{tab:stats} shows statistics of \prop. As can be seen, the training set is small, which makes learning challenging.

Along with the dataset, \citet{dalvi2018tracking} introduce the task of tracking state changes at a fine-grained sentence level. To solve this task, a model must answer three categories of questions (10 questions in total) about an entity $E$: (1) Is $E$ created, (destroyed, moved) in the process? (2) When (step \#) is $E$ created, (destroyed, moved)? (3) Where is $E$ created, (destroyed, moved from/to)? Cat. 1 asks boolean questions about the existence and movement of entities. Cat. 2 and 3 are harder tasks, as the model must correctly predict the step number at which a state changes as well as the correct locations (text spans) of entities at each step.

\citet{tandon2018reasoning} introduce a second task on the \prop dataset that measures state changes at a coarser \emph{process} level. To solve this task, a model must correctly answer the following four types of questions: (1) What are the inputs to the process? (2) What are the outputs of the process? (3) What conversions occur, when and where? (4) What movements occur, when and where? Inputs to a process are defined as entities that exist at the start of the process but not at the end and outputs are entities that exist at the end of the process and were created during it. A conversion is when some entities are created and others destroyed, while movements refer to changes in location. \citet{dalvi2018tracking} and \citet{tandon2018reasoning} propose different models to solve each of these tasks separately, whereas we evaluate the same model, \alg, on both tasks.

\citet{bosselut2017recipe} recently released the \recipes dataset, which has various annotated states (e.g. shape, composition, location, etc.) for ingredients in cooking recipes. We further test \alg on the location task to align with our \prop experiments. This is arguably the dataset's hardest task, since it requires classification over more than 260 classes while the others have a much smaller label space (maximum of 4).
Note that rather than treating this problem as classification over a fixed lexicon as in previous models, our model aims to find the location-describing span of text in the recipe paragraph.

\section{Model}
\label{sec:model}
\alg tracks the temporal state change of entities in procedural text. Naturally, the model is entity-centric \citep{henaff2016tracking,bansal2017relnet}: it associates each participant entity of the procedural text with a unique node and embedding in its internal graph. \alg is also equipped with a neural machine reading comprehension model which is queried about the current location of each entity.

At a high level, our model operates as follows. We summarize some important notation in Table~\ref{tab:my_label}.
\alg processes a paragraph $p = \lbrace w_j \rbrace^P_{j=1}$, of $P$ words, by incrementally reading \emph{prefixes} of the paragraph up to and including sentence $s_t$ at each time step $t$. This continues until it has seen all sentences $\lbrace s_t \rbrace^T_{t=1}$ of the paragraph.
At each time step (sentence) $t$, we engage the MRC module to query for the state of each participant entity (participants are known in \prop \textit{a priori}).
The query process conditions on both the input text and the target entity's node embedding, $e_{i,t-1}$, where the latter comes from the graph at the previous time step.
The MRC module returns an answer span describing the entity's current location at $t$; we encode this text span as the vector $\psi_{i,t}$.
Conditioning on the span vectors $\psi_{i,t}$, the model constructs the graph $G_t$ by updating $G_{t-1}$ from the previous time step.

The model's knowledge graphs $G_t$ are bipartite, having two sets of nodes with implied connections between them: $G_t = \{e_{i,t}, \lambda_{i,t} \}$.
Each node denotes either an entity ($e_{i,t}$) or that entity's corresponding location ($\lambda_{i,t}$), and is associated with a real-valued vector.
We use $e_{i,t}$ and $\lambda_{i,t}$ to denote nodes in the graph \emph{and} their vector representations interchangeably.
The bipartite graphs $G_t$ have only one (implicit) relation type, the current location, though we plan to extend this in future work.
To derive $G_t$ from its previous iterate $G_{t-1}$, we combine both hard and soft graph updates.
The update to an entity's node representation with new location information arises from a hard decision made by the MRC model, whereas co-reference between entities across time steps is resolved with soft attention.
We now describe all components of the model in detail.

\subsection{Entity and Span Representations}
\label{sub:context_rnns}
In the \prop dataset, entities appear in the paragraph text.\footnote{We compute the positions of the occurrence of entities by simple string matching.}
Therefore, we \emph{derive} the initial entity representations from contextualized hidden vectors by encoding the paragraph with a bi-directional LSTM \citep{hochreiter1997long}. This choice has the added advantage that initial entity representations share information through context, unlike in previous models~\citep{henaff2016tracking,das2017question,bansal2017relnet}.
Entities in the dataset can be multi-word expressions (e.g., \textit{electric oven}). To obtain a single representation, we concatenate the contextualized hidden vectors corresponding to the start and end span tokens and take a linear projection.~i.e., if the mention of entity $i$ occurs between the $j$-th and $j+k$-th position, then the initial entity representation $\nu_{i}$ is computed as $\nu_{i} = W_e [c_{j};c_{j+k}] + b_e.$
We use $i$ to index an entity and its corresponding location, while $c_{j}$ represents the contextualized hidden vectors for token $j$ and $[;]$ represents the concatenate operation. An entity may occur multiple times within a paragraph. We give equal importance to all occurrences by summing the representations for each.

When queried about the current location of an entity, the MRC module (\S~\ref{sub:dynamic_mrc}) returns a span of text as the answer, whose representation is later used to update the appropriate node vector in the graph. We obtain this answer-span representation analogously as above, and denote it with $\psi_{i,t}$.

\subsection{Machine Reading Comprehension Model}
\label{sub:dynamic_mrc}
Rather than design a specialized MRC architecture, we make simple extensions to a widely used model -- \drqa \citep{chen2017reading} --
to adapt it to query about the evolving states of entities.
In summary, our modified \drqa implementation operates on prefixes of sentences rather than the full paragraph (like \proglobal), and at each sentence (time step) it conditions on both the current sentence representation $s_t$ and the dynamic entity representations in $G_{t-1}$.

For complete details of the \drqa model, we refer readers to the original publication \citep{chen2017reading}.
Broadly, it uses a multi-layer recurrent neural network (RNN) architecture for encoding both the passage and question text and uses self-attention to match these two encodings. For each token $j$ in the text, it outputs a score indicating its likelihood of being the start or end of the span that answers the question. We reuse all of these operations in our model, modified as described below.

\begin{table}[t]
    \centering
    \footnotesize
    \begin{tabular}{l l}\hline
       Notation  & Meaning \\\hline
        $N \in \mathbb{N}$ & Number of participant entities in the process.\\
        $\nu_{i} \in \mathbb{R}^{d}$ & Initial entity representation, derived from the text, for the $i$-th entity at time $t$ = 0 (\S~\ref{sub:context_rnns}) \\
        $e_{i,t} \in \mathbb{R}^{d}$ & Entity node representation for the $i$-th entity at time $t$, in the graph $G_t$ (\S~\ref{sub:graph_update}) \\
        $\psi_{i,t} \in \mathbb{R}^{d}$ & Location representation derived from the text for the $i$-th entity at time $t$ (\S~\ref{sub:context_rnns}) \\
        $\lambda_{i,t} \in \mathbb{R}^{d}$ & Location node representation for the $i$-th entity at time $t$, in the graph $G_t$ (\S~\ref{sub:coref},~\ref{sub:graph_update})\\
        $\Lambda_{t} \in \mathbb{R}^{N \times d}$ & Matrix of all location node representations at time $t$ \\
        $U_{t} \in \mathbb{R}^{N \times N}$ & Soft co-reference matrix at time step $t$ (\S~\ref{sub:coref})\\\hline
    \end{tabular}
    \caption{Symbols used in Section~\ref{sec:model}. The text-based representations of entities and locations are derived from the hidden representations of the context-RNN (\S~\ref{sub:context_rnns}). The node representations are added to the graph $G_t$ at the end of time step $t$ (\S~\ref{sub:graph_update}).}
    \label{tab:my_label}
\end{table}
We query the \drqa model about the state of each participant entity at each time step $t$. This involves reading the paragraph up to and including sentence $s_t$. To query, we generate simple natural language questions for an entity, $E$, such as ``Where is $E$ located?'' This is motivated by the work of \citet{Levy2017ZeroShotRE}.
Our \drqa component also conditions on entities.
Recall that vector $e_{i,t-1}$ denotes the entity's representation in the knowledge graph $G_{t-1}$.
The module conditions on $e_{i,t-1}$ in its output layer, basically the same way as the question representation is used in the \emph{output alignment} step in \citet{chen2017reading}.
However, instead of taking a bi-linear map between the question and passage representations as in that work, we first concatenate the question representation with $e_{i,t-1}$ and pass the concatenation through a 2-layer MLP.
This yields an entity-dependent question representation.
We use this to compute the output start and end scores for each token position, taking the $\arg\max$ to obtain the most likely span.
As mentioned, we encode this span as vector $\psi_{i,t}$ (\S~\ref{sub:context_rnns}).

The \prop dataset includes two special locations that don't appear as text spans: \textit{nowhere} and \textit{somewhere}. The current location of an entity is \textit{nowhere} when the entity does not exist yet or has been destroyed, whereas it is \textit{somewhere} when the entity exists but its location is unknown from the text. 
Since these locations don't appear as tokens in the text, the span-predictive MRC module cannot extract them. Following \citet{dalvi2018tracking}, we address this with a separate classifier that predicts, given a graph entity node and the text, whether the entity represented by the node is \textit{nowhere}, \textit{somewhere}, or its location is stated. We learn the location-node representations for \textit{nowhere} and \textit{somewhere} during training.

\subsection{Soft Co-reference}
\label{sub:coref}
\begin{figure*}
    \centering
    \vspace{-2em}
    \includegraphics[width=0.65\columnwidth]{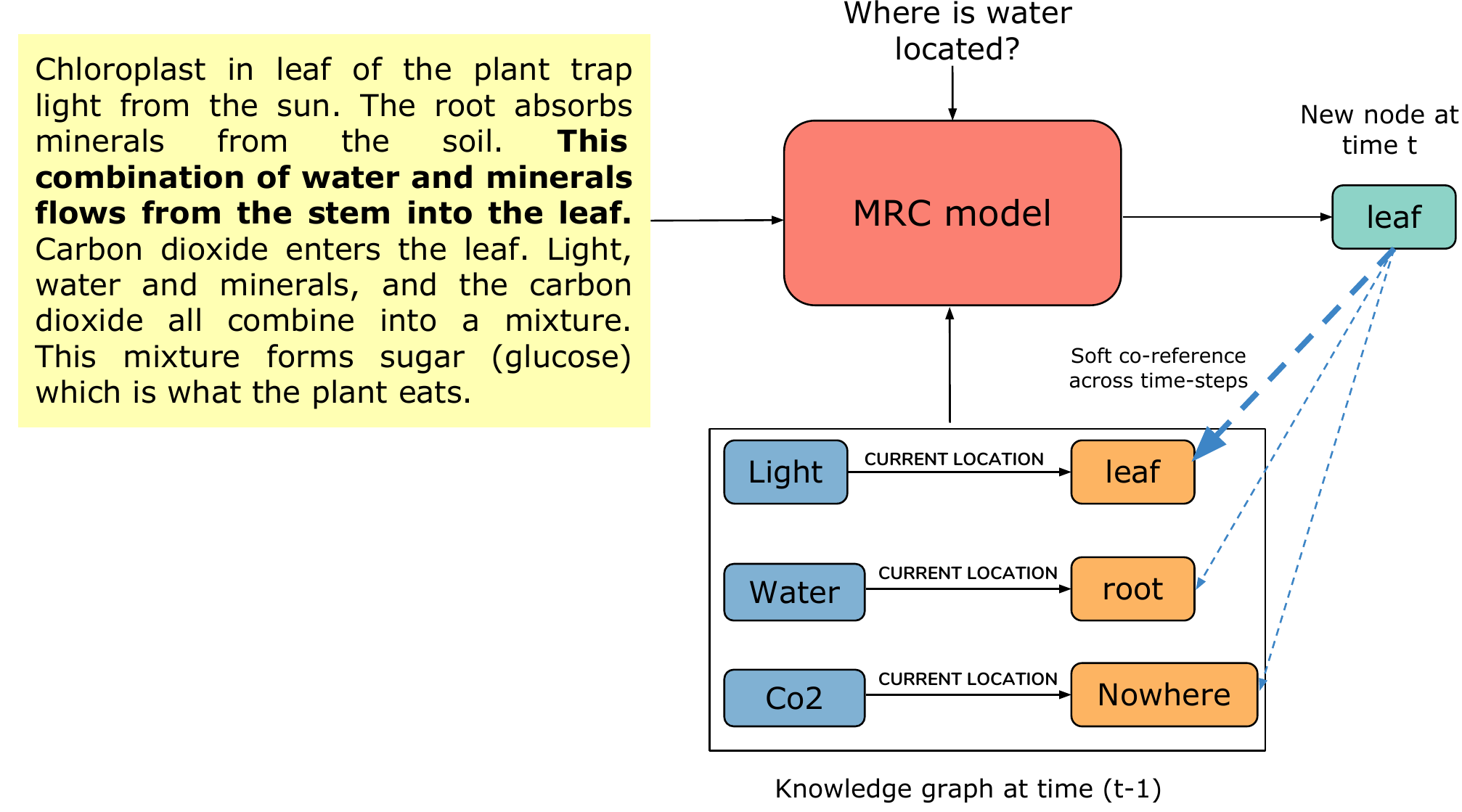}
    \caption{Soft co-reference across time steps. The sentence at the current time step is highlighted. When the MRC model predicts a span (\textit{leaf}) present in the graph at the previous time step, \alg does soft attention and a gated update to preserve information across time steps (\S~\ref{sub:coref}). The thicker arrow shows higher attention weight between the old and new node.}
    \label{fig:soft_coref}
\end{figure*}

\ignore{
It is also possible that the state of few entities \emph{do not} change at a time step. For example, consider the sentence (in bold) in figure~\ref{fig:soft_coref}. At this time-step (sentence), only the location of \textit{water} and \textit{minerals} change (to \textit{leaf}) and the location of \textit{light} remains unchanged (as \textit{leaf}). Since there already exists a node corresponding to \textit{leaf} in the graph, instead of replacing the current \textit{leaf} node, our model performs `\textit{soft co-reference}' (\S\ref{sub:coref}) by attending to the locations of entities at the previous time step. Co-reference with soft attention \citep{bahdanau2014neural,luong2015effective} also provides other advantages. For example, the reading comprehension model can return different span for the same surface form (e.g. different mentions of \textit{leaf} in the first and third sentence). Soft attention could still identify that they are the same spans by learning high attention weights between them. It is also robust when the MRC model returns similar spans (e.g. \textit{the leaf} instead of \textit{leaf}).
Since a node can already exist in the graph $G_{t-1}$, we perform a soft co-reference resolution between the initial node vector $v_{i,t}$ and the existing nodes $R_{t-1} = \lbrace r^{L}_{i,t-1} \rbrace^N_{i=1}$ from the previous time step.
}

To handle cases when entity states do not change and when states are referred to with different surface forms (either of which could lead to undesired node duplication), our model uses soft co-reference mechanisms (Figure~\ref{fig:soft_coref}) both across and within time steps. 
Disambiguation across time steps is accomplished by attention and a gated update, using the incoming location vector $\psi_{i,t}$ and the location node representations from the previous time step:
\begin{equation}
\begin{aligned}
\small
& a_{i,t} = \text{softmax}(\Lambda_{t-1} \psi_{i,t}) \\
& \psi'_{i, t} =  \Lambda_{t-1}^{\top} a_{i,t} \\
& g_{i} = \text{sigmoid}(W_{i} [\psi'_{i,t}; \psi_{i,t}] + b_{i}) \\
& \lambda'_{i,t}  = g_{i} \psi_{i,t} + (1 - g_{i}) \psi'_{i,t},
\label{eq:coref}
\end{aligned}
\end{equation}
where $\Lambda_{t-1} = [ \lambda_{i,t} ]^N_{i=1} \in \mathbb{R}^{N \times d}$ is a matrix of location node representations from the previous time step (stacked row-wise) and  $\psi_{i,t}$ is the location span vector output by the MRC module.
The result vector $\lambda'_{i,t}$ is a disambiguated intermediate node representation.

This process only partially addresses node de-duplication. Since different instances of the same location can be predicted for multiple entities, we also perform a co-reference disambiguation within each time step using a self-attention mechanism:
\begin{equation}
\begin{aligned}
\small
& u_{i,t} = \text{softmax}(\Lambda'_{t} \lambda'_{i,t}) \\
& \lambda_{i,t} =  \Lambda'^{\top}_{t} u_{i,t},
\label{eq:coref2}
\end{aligned}
\end{equation}
where $\Lambda'_{t} = [ \lambda'_{i,t} ]^N_{i=1} \in \mathbb{R}^{N \times d}$ is a matrix of intermediate node representations (stacked row-wise) and $U_{t} = [ u_{i,t} ]^N_{i=1} \in \mathbb{R}^{N \times N}$ is a co-reference adjacency matrix. We calculate this adjacency matrix at the beginning of each time step to track related nodes within $t$, and re-use it in the graph update step.

\subsection{Graph Update}
\label{sub:graph_update}

The graph update proceeds according to the following set of equations for each update layer $l$:
\begin{equation}
\begin{aligned}
\small
h^{l}_{i,t} &= \text{LSTM}([e^{l-1}_{i,t};\lambda^{l-1}_{i,t}; h^{l}_{i,t-1}]) \\
e^{l}_{i,t} &= e^{l-1}_{i,t} + h^{l}_{i,t} \\
\tilde{\lambda}^{l}_{i,t} &= \lambda^{l-1}_{i,t} + h^{l}_{i,t} \\
\lambda^{l}_{i,t} &=  \tilde{\Lambda}^{l}_{t} u_{i,t}.
\end{aligned}
\end{equation}
We first compose all connected entity and location nodes with their history summary, $h^{l}_{i,t-1}$, using an LSTM unit.
Next, the updated node information is attached to the entity and location representations through two residual updates \citep{he2016deep}.
These propagate information between the entity and location representations; i.e., if two entities are at the same location, then the corresponding entity representations will receive a similar update. Likewise, location representations are updated with pertinent entity information.
Last, we perform a co-reference pooling operation for the location node representations. This uses the adjacency matrix $U_t$, where $\tilde{\Lambda}^{l}_{t}$ is a row-wise stacked matrix of the $\tilde{\lambda}^{l}_{i,t}$,
to tie co-referent location nodes together.

The recurrent graph module stacks $L$ such layers to propagate node information along the graph's edges. The resulting node representations are $e^{L}_{i,t}$ and $\lambda^{L}_{i,t}$ for each participant entity and its location. We use $e_{i,t} = e^{L}_{i,t}$ to condition the MRC model, as described in \S\ref{sub:dynamic_mrc}.
We make use of this particular graph module structure, rather than adopting an existing model like GraphCNNs~\citep{graphcnns,kipf2016semi}, because recurrent networks are designed to propagate information through time.

\subsection{Training}
\label{sub:train}
The full \alg model is trained end-to-end by minimizing the negative log-likelihood of the correct span tokens under the MRC module's output distribution and the textual entailment model. This is a fairly soft supervision signal, since we do not train the graph construction modules directly.
We teacher-force the model at training time by updating the location-node representations with the encoding of the correct span.
We do not pretrain the MRC module, but we represent paragraph tokens with pretrained FastText embeddings~\citep{joulin2016fasttext}.
See the appendix~\ref{sec:implement_detail} for full implementation and training details.

\section{Experiments and Discussion}
\label{sec:experiments}
\vspace{-2mm}
We evaluate our model on three different tasks. We also provide an ablation study along with quantitative and qualitative analyses to highlight the performance contributions of each module.
\vspace{-6mm}
\subsection{Results on Procedural Text}
\vspace{-2mm}
We benchmarked our model on two \prop comprehension tasks introduced respectively in \citet{dalvi2018tracking} and \citet{tandon2018reasoning}.
Refer to Section~\ref{sec:data} for a detailed description about the data and tasks.
\citet{dalvi2018tracking} and \citet{tandon2018reasoning} respectively introduce a specific model for each task, whereas we test \alg on both tasks. A primary motivation for building KGs is because they can be \emph{queried} for salient knowledge in downstream applications. We evaluate \alg on the above two tasks by querying the KGs it builds at each time-step; we use the official evaluation pipeline\footnote{\url{https://github.com/allenai/propara/tree/master/propara/eval}} for each task. In results below, we report an average score of three runs of our model with different hyperparameter settings.

\subsubsection{Task 1: Sentence-level Evaluation}
\vspace{-1mm}
Table ~\ref{tab:naacl} shows our main results on the first task. Following the original task evaluation, we report model accuracy on each subtask category and macro and micro averages over the subtasks. 

Human performance is 79.69\%, micro-average. A state-of-the-art memory augmented network, \entnet \citep{henaff2016tracking}, which is built to track entities but lacks an explicit graph structure, achieves 25.96\%.
The previous best performing model is \proglobal, which achieves 45.37\%. Our \alg improves over this result by 1.25\% absolute score in terms of micro-averaged accuracy. Comparing various models for each subtask category, \proglobal leads in Category 1 by a small margin of around 0.1\%. For the more challenging Categories 2 and 3, \alg outperforms \proglobal by a large margin. These questions require fine-grained predictions of state changes.

\begin{table}[H]
    \centering
    \scriptsize
    \begin{tabular}{r|c|c|c|c|c}
        \toprule
        & Cat 1 & Cat 2 & Cat 3 & Macro-avg & Micro-avg \\
        \midrule
        \text{Human upper bound}                     & 91.67 & 87.66 & 62.96 & 80.76 & 79.69 \\
        \midrule
        \text{Majority}                              & 51.01 & -- & -- & -- & -- \\
        \text{Rule based}                            & 57.14 & 20.33 & 2.40 & 26.62 & 26.24 \\
        \text{Feature based}                         & 58.64 & 20.82 & 9.66 & 29.7 & 29.64 \\
        \text{EntNet (\cite{henaff2016tracking})}    & 51.62 & 18.83 & 7.77 & 26.07 & 25.96 \\
        \text{Pro-Local (\cite{dalvi2018tracking})}  & 62.65 & 30.50 & 10.35 & 34.50 & 33.96 \\
        \text{Pro-Global (\cite{dalvi2018tracking})} & \textbf{62.95} & 36.39 & 35.90 & 45.08 & 45.37 \\
        \text{\alg (ours)}                                  & 62.86 & \textbf{40.00} & \textbf{38.23} & \textbf{47.03} & \textbf{46.62} \\
        \bottomrule
    \end{tabular}
    \caption{Task 1 results (accuracy).}
    \label{tab:naacl}
\end{table}
\vspace{-6mm}
\subsubsection{Task 2: Document-level Evaluation}

We report the performance of our model on the document-level task, along with previously published results, in Table~\ref{tab:emnlp}. The same \alg model achieves 3.02\% absolute improvement in $\mathrm{F_{1}}$ over the previous best result of \prostruct. \prostruct incorporates a set of commonsense constraints for globally consistent predictions. We analyzed \alg's outputs and were surprised to discover that our model learns these commonsense constraints from the data in an end-to-end fashion, as we show quantitatively in \S\ref{sub:cs_constraints}.
\vspace{-2mm}
\begin{table}[H]
    \centering
    \scriptsize
    \begin{tabular}{r|c|c|c}
        \toprule
        & Precision & Recall & $\mathrm{F_{1}}$ \\
        \midrule
        \text{Pro-Local (\cite{dalvi2018tracking})}  & \textbf{77.4} & 22.9 & 35.3 \\
        \text{QRN (\cite{seo2016query})}                         & 55.5 & 31.3 & 40.0 \\
        \text{EntNet (\cite{henaff2016tracking})}    & 50.2 & 33.5 & 40.2 \\
        \text{Pro-Global (\cite{dalvi2018tracking})} & 46.7 & \textbf{52.4} & 49.4 \\
        \text{Pro-Struct (\cite{tandon2018reasoning})} & 74.2 & 42.1 & 53.75 \\
        \text{\alg (ours)}                                  & 64.52 & 50.68 & \textbf{56.77} \\
        \bottomrule
    \end{tabular}
    \caption{Task 2 results.}
    \label{tab:emnlp}
\end{table}
\vspace{-8mm}

\subsection{Recipe Description Experiments}
\vspace{-1mm}
\ignore{
\begin{table}[H]
    \centering
    \scriptsize
    \begin{tabular}{r|c}
        \toprule
        & $\mathrm{F_{1}}$ \\
        \midrule
        \text{Neural Process Networks (\cite{bosselut2017recipe})}  & 51.28 \\
        \text{\alg} (10K training samples) & \textbf{51.64} \\
        \bottomrule
    \end{tabular}
    \caption{Experimental results on the \recipes dataset.}
    \label{tab:recipe}
\end{table}
}

We also evaluate our model on the \recipes dataset, where we predict the locations of cooking ingredients. In the original work of \citet{bosselut2017recipe}, they treat this problem as classification over a fixed lexicon of locations, whereas \alg searches for the correct location span in the text. Our model slightly outperforms the baseline NPN model on this task even after it was trained on just 10K examples (the full training set is around 60K examples): NPN achieves 51.28\% $\mathrm{F_{1}}$ training on all the data, while \alg achieves \textbf{51.64}\% $\mathrm{F_{1}}$ after 10k examples.


\subsection{Ablation Study}
\vspace{-2mm}
We performed an ablation study to evaluate different model variations on \prop Task 1. The main results are reported in Table~\ref{tab:ablation}. Removing the soft co-reference disambiguation within time steps (Equations~\ref{eq:coref2}) from \alg resulted in around 1\% performance drop. The drop is more significant when the co-reference disambiguation across time steps (Equations~\ref{eq:coref}) is removed.

We also replaced the recurrent graph module with the standard LSTM unit and used the LSTM hidden state for the entity representation. As this model variation lacks the information propagation across graph nodes, we observed a large performance decrease.

For the last two variations, we simply train the MRC model in isolation and predict location spans from the current sentence or paragraph prefix text (i.e., the current and all previous sentences). These models construct no internal knowledge graphs. We can see that training the MRC model on paragraph prefixes already provides a good starting performance of 40.83\% micro-average, which is significantly boosted by the recurrent graph module and graph conditioning up to 47.64\%.

\begin{table}[H]
    \centering
    \scriptsize
    \begin{tabular}{r|c|c|c|c|c}
        \toprule
        & Cat 1 & Cat 2 & Cat 3 & Macro-avg & Micro-avg \\
        \midrule
        \text{\alg} & 58.55 & 38.52 & 42.22 & 46.43 & 47.64 \\
        \text{- Coref across time steps} & 61.07 & 37.38 & 35.58 & 44.68 & 46.32 \\
        \text{- Coref within time step} & 57.88 & 38.09 & 40.19 & 45.39 & 46.63 \\
        \midrule
        \text{Standard LSTM as graph unit} & 56.84 & 13.15 & 10.95 & 26.98 & 29.97 \\
        
        \text{MRC on entire paragraph} & 58.85 & 21.82 & 26.52 & 35.73 & 35.98 \\
        \text{MRC on prefix} & 61.28 & 32.58 & 29.48 & 41.11 & 40.83 \\
        \bottomrule
    \end{tabular}
    \caption{Ablation experiment results}
    \label{tab:ablation}
\end{table}

\subsection{Commonsense Constraints}
\label{sub:cs_constraints}

For accurate, globally consistent predictions for the second task, \cite{tandon2018reasoning} introduced a set of commonsense constraints on their model. Stated in natural language, these constraints are: 1) An entity must \textbf{exist} before it can be \textbf{moved} or \textbf{destroyed}; 2) An entity cannot be \textbf{created} if it already \textbf{exists}; 3) An entity cannot \textbf{change} until it is \textbf{mentioned} in the paragraph.

To analyze whether our model can learn the above constraints from data, we count the number of predictions that violate any constraints on the test set. In Table~\ref{tab:constraints} we compare the behavior of different models by computing the number of violations made by \cite{tandon2018reasoning}'s model and several variants of our model. Note that we only count instances where a model predicts an entity state change.

\begin{table}[H]
    \centering
    \scriptsize
    \begin{tabular}{r|c|c|c}
        \toprule
        Model & State Change Predictions & Violations & Violation Proportion (\%) \\
        \midrule
        \text{Pro-Struct (\cite{tandon2018reasoning})} & 270 & 17 & 6.30 \\
        \text{MRC on entire paragraph} & 381 & 104 & 27.30 \\
        \text{MRC on prefix} & 703 & 154 & 21.93 \\
        \text{Standard LSTM as graph unit} & 447 & 20 & 4.47 \\
        \text{\alg} & 466 & 19 & \textbf{4.08} \\
        \bottomrule
    \end{tabular}
    \caption{Commonsense constraints violation.}
    \label{tab:constraints}
\end{table}

To our surprise, \alg learns to violate fewer constraints (proportionally) than \prostruct even without explicitly training it to do so. As the table shows, MRC models without recurrent graph modules perform worse in terms of constraint violations than both \alg and a model with standard LSTM as its graph unit. This suggests that recurrency and graph representations play an important role in helping the model to learn commonsense constraints.

\subsection{Qualitative Analysis}
\vspace{-2mm}
We picked an example from the test data and took a closer look at the model outputs to investigate how \alg dynamically adjusts its decisions via the dynamic graph module and finds accurate spans with the conditional MRC model. The step-by-step output of both \proglobal (\cite{dalvi2018tracking}) and \alg is shown in Table~\ref{tab:cherry_pick}, where we track the state of entity \textit{blood} across six sentences. \alg outputs smoother and more accurate predictions.
\begin{table}[H]
    \centering
    \scriptsize
    \begin{tabular}{l|c|c}
        \toprule
        Sentences & \multicolumn{2}{c}{Location of entities after each sentence} \\
        \midrule
        \text{(Before first sentence)} & \text{\textcolor{orange}{somewhere}} & \text{\textcolor{red}{somewhere}}\\
        \text{Blood enters the \textcolor{blue}{right side of your heart}.} & \text{\textcolor{orange}{heart}} & \text{\textcolor{red}{right side of your heart}} \\
        \text{Blood travels to the \textcolor{blue}{lungs}.} & \text{\textcolor{orange}{lung}} & \text{\textcolor{red}{lungs}} \\
        \text{Carbon dioxide is removed from the blood.} & \text{\textcolor{orange}{blood}} & \text{\textcolor{red}{lungs}} \\
        \text{Oxygen is added to your blood.} & \text{\textcolor{orange}{lung}} & \text{\textcolor{red}{lungs}} \\
        \text{Blood returns to \textcolor{blue}{left side of your heart}.} & \text{\textcolor{orange}{blood}} & \text{\textcolor{red}{heart}} \\
        \text{The blood travels through the \textcolor{blue}{body}.} & \text{\textcolor{orange}{body}} & \text{\textcolor{red}{body}} \\
        \bottomrule
    \end{tabular}
    \caption{Two models' predictions of entity locations, on randomly selected paragraph about blood circulation. In this example the entity is \textbf{blood}. Predicted results from Pro-Local (\cite{dalvi2018tracking}) are in \textcolor{orange}{orange}, results from \alg are in \textcolor{red}{red}, important locations in paragraph are in \textcolor{blue}{blue}.}
    \label{tab:cherry_pick}
\end{table}

\section{Conclusion}
\label{sec:conc}
We proposed a neural machine-reading model that constructs dynamic knowledge
graphs from text to track locations of participant entities in procedural text.
It further uses these graphical representations to improve its downstream comprehension of text.
Our model, \alg, achieves state-of-the-art results on two question-answering tasks from the \prop dataset and one from the \recipes dataset.
In future work, we will extend the model to construct more general knowledge graphs with multiple relation types.

\bibliography{iclr2019_conference}
\bibliographystyle{iclr2019_conference}
\clearpage
\appendix
\section{Implementation Details}
\label{sec:implement_detail}

Implementation details of \alg are as follows. \\
In all experiments, the word embeddings are initialized with FastText embeddings \citep{joulin2016fasttext}; we use a document LSTM with two layers, the number of hidden units in each layer is 64. We apply dropout rate of 0.4 in all recurrent layers, and 0.3 in all other layers. The number of recurrent graph layers were set to ($L=2$). The hidden unit size for the recurrent graph component was set to 64.

During training, the mini-batch size is 8. We use \emph{adam} \citep{kingma2014adam} as the step rule for optimization, The learning rate is set to 0.002. The model is implemented using \textit{PyTorch} \citep{paszke2017automatic}.

\end{document}